\documentclass[runningheads]{llncs}
\usepackage[T1]{fontenc}
\usepackage{tablefootnote}
\usepackage[labelfont=bf,size=small]{caption} % for customizing caption styles
\usepackage{amsfonts}  % for \mathcal and \mathfrak
\usepackage[table]{xcolor}

\usepackage{multirow}
\usepackage{graphicx}
\usepackage{array}
\usepackage{caption}
\usepackage{wrapfig}
\usepackage{tabularray}
\usepackage{amsmath}
\usepackage{amssymb}
\usepackage{threeparttable}
\usepackage{booktabs}
\usepackage{graphicx}
\usepackage{multirow}
\usepackage{multicol}
\usepackage{tabularx}
\usepackage{subcaption}

\usepackage[margin=1.35in]{geometry}

\usepackage{xcolor}
\usepackage{tabularx}
\usepackage{colortbl}
\usepackage{orcidlink}

\definecolor{lime}{rgb}{0.88,2,10}

\usepackage[spaces,hyphens]{xurl}
\newcommand*{\Resize}[2]{\resizebox{#1}{!}{$#2$}}%

\newcommand{\fref}[1]{Fig.~\ref{#1}}
\newcommand{\tref}[1]{Table~\ref{#1}}

\newcommand\HUGE{\fontsize{13.2}{20}\selectfont}

\usepackage{fancyhdr}
\fancypagestyle{mystyle}{
    \chead{28$^{th}$ Pacific-Asia Conference on Knowledge Discovery and Data Mining (PAKDD 2024)}}

\begin{document}

\title{\HUGE Unmasking Dementia Detection by Masking Input Gradients: \\A JSM Approach to Model Interpretability and Precision}\vspace{-1cm}

%JSM-based Model Debugging
%\title{What Your Loss Function Won't Tell You}

\author{Yasmine Mustafa\orcidlink{0009-0002-3512-9659} 
\and
Tie Luo\thanks{Corresponding author.}\orcidlink{0000-0003-2947-3111}
}

\institute{CS Department, Missouri University of Science and Technology, Rolla, MO 65409, USA
\email{\{yam64,tluo\}@mst.edu}}
 \maketitle  
\thispagestyle{mystyle}

\vspace{-2mm}

\begin{abstract}

The evolution of deep learning and artificial intelligence has significantly reshaped technological landscapes. However, their effective application in crucial sectors such as medicine demands more than just superior performance, but trustworthiness as well. While interpretability plays a pivotal role, existing explainable AI (XAI) approaches often do not reveal \textit{Clever Hans} behavior where a model makes (ungeneralizable) correct predictions using spurious correlations or biases in data. Likewise, current post-hoc XAI methods are susceptible to generating unjustified counterfactual examples. In this paper, we approach XAI with an innovative {\em model debugging} methodology realized through Jacobian Saliency Map (JSM). To cast the problem into a concrete context, we employ Alzheimer's disease (AD) diagnosis as the use case, motivated by its significant impact on human lives and the formidable challenge in its early detection, stemming from the intricate nature of its progression. 
We introduce an interpretable, multimodal model for AD classification over its multi-stage progression, incorporating JSM as a modality-agnostic tool that provides insights into volumetric changes indicative of brain abnormalities. Our extensive evaluation including ablation study manifests the efficacy of using JSM for model debugging and interpretation, while significantly enhancing model accuracy as well.

\keywords{Trustworthy AI \and Interpretability \and Explainability \and Reliability \and Jacobian saliency map \and Alzheimer's disease}
\end{abstract}

\section{Introduction}
Despite the remarkable successes of deep learning and artificial intelligence across various technological domains, achieving a high-performing system does not automatically guarantee its practical deployment and use, particularly in the field of medicine. Given the profound implications of medical decisions on human lives, doctors and patients often approach AI diagnoses with skepticism, notwithstanding claims of high precision, due to concerns surrounding trustworthiness.

In this study, we delve into the concept of trustworthy medical AI, focusing on two essential aspects. Firstly, \textit{explainability} is crucial; without a clear explanation of the rationale behind a diagnosis, patients would be much less receptive to AI-driven decisions. Secondly, \textit{reliability} is paramount, ensuring that AI models make predictions based on pertaining patterns rather than exhibiting what is known as ``Clever-Hans behavior.'' This phenomenon occurs when a machine learning model seemingly performs well but makes decisions based on irrelevant factors such as biases or coincidental correlations in data.

These two aspects are interrelated: a thorough model explanation not only provides the foundation for decision-making but also reveals whether predictions are influenced by Clever-Hans behavior. By addressing both explainability and reliability, we aim to enhance trust in medical AI systems and pave the way for their responsible and effective integration into healthcare practices.

A wide variety of explainable AI (XAI) tools have arisen to explain the predictions of trained \textit{black-box} models, referred to as \textit{post-hoc methods}. Although these methods have shown some promising prospects, they are vulnerable to the risk of generating unjustified counterfactual examples \cite{laugel2019dangers}, hence may not be reliable. It is worth noting the nuances between \textit{explainability} and \textit{interpretability} in this context, although they are often loosely used interchangeably. \textit{Explainability} refers to the ability to explain model decisions after training (post-hoc), while the original model is not interpretable by itself. On the other hand, \textit{interpretability} is an inherent property of a model and means how easily and intuitively one can understand and make sense of the model's decision-making process. Examples of highly interpretable models include decision trees and linear regression, which provide easily traceable logic for what roles the features played in decision-making. However, such models are \textit{shallow models} and typically under-perform deep neural networks (DNNs). As post-hoc explanation is often inadequate to unravel the full complexity of model behavior due to its after-training nature, we focus on designing \textit{interpretable} models in this paper, taking a {\em during-modeling} approach.

In this paper, we introduce a novel approach that guides the decision-making process of neural networks during the training phase by not only directing the model toward correct predictions but also penalizing any (including correct) predictions based on \textit{wrong cues}. Cues refer to patterns, relationships, or features within the input data that are deemed relevant to the task at hand by the model to make predictions or decisions. Misinterpreted or misidentified cues erode the trustworthiness and reliability of a model even if its predictions are correct since the performance would not generalize to future unseen data. 

For a concrete problem context, we take Alzheimer's disease (AD) as the specific medical condition in this study, but highlight that our approach can be extended to similar problems without change of principle. AD is the predominant form of dementia and a major contributor to mortality. It impacts brain areas that are responsible for thought, memory, and language and is hard to cure. Although changes in the brain can manifest long before AD symptoms appear, it is challenging for medical professionals to detect manually until AD reaches late and severe stages, which however are no longer reversible. Therefore, AI-based diagnosis of AD has been researched actively. However, to date, newly developed models as such have rarely been adopted in clinical decision support systems (CDSS) for primary care, because such new models are almost exclusively based on DNNs which lack explainability.

Another crucial aspect of AD is that its clinical representation often involves multiple modalities such as computed tomography (CT) and magnetic resonance imaging (MRI) images. While combining such data modalities could lead to better precision \cite{venugopalan2021multimodal}, it poses a further challenge in designing interpretable models to interpret decisions involving all modalities.

In this paper, we propose an interpretable, multimodal model for classifying AD across its multi-stage progression, including early detection. We introduce a \textit{Jacobian-Augmented Loss} function (JAL) that incorporates Jacobian Saliency Maps (JSM) as a model {\em self-debugger}. This approach is \textit{modality-agnostic} as it is not limited to any specific imaging modality but can work seamlessly with various integrated modalities, making it versatile and adaptable to different data sources. In this study, we focus on images, given that other modalities such as Mini-Mental State Examination (MMSE) can be diagnosed effectively by medical professionals, and image-based early detection of AD proves challenging.
Our methodology aligns with the broader concept of \textit{model debugging} that involves troubleshooting unwanted behaviors and examining models' predictions. Our goal is to ensure that a model not only learns to make accurate predictions but also avoids wrong cues as in the \textit{Clever Hans} phenomenon, thus enhancing model reliability. We achieve this using JSM, which is computed during image preprocessing. Besides debugging, JSM also allows us to enhance model precision by highlighting deformations in body issues (e.g., brain as in the context of AD).

In summary, this paper makes the following contributions:
\begin{itemize}
    \item We design a novel loss function JAL which incorporates Jacobian saliency maps (JSM) to enable machine learning models to self-debug its decision-making process automatically. This approach ensures the model predictions to be based on genuine patterns and cues, and renders the model decision-making process to be more interpretable through a during-modeling methodology.
    \item We include multimodal data fusion into the process through two distinct fusion techniques, not only shedding light on their differences but also showcasing the adaptability of JAL to different modalities and fusion levels.
    \item Our approach enables models to provide fine-grained classification in terms of 4 classes including cognitively normal (CN) and three main AD stages: mild cognitive impairment (MCI), mild AD, and moderate to severe AD. On the contrary, existing approaches only provide binary classification or combine two or more stages into one class. Furthermore, although coarse-grained approaches are {\em easier} to attain higher accuracy due to less classes, our fine-grained diagnosis achieves higher accuracy than them.

    \item Our comprehensive evaluation including ablation study proves the efficacy of using JSM as a model self-debugger for producing both reliable predictions and trustworthy interpretations. 
\end{itemize}

\section{Related Work}
El-sappagh et al. \cite{el2021multilayer}'s primary goal in explainability is to provide post-hoc explanations for the decisions made by Random Forest (RF) on classifying AD based on multi-modal data. Hence, the method does not try to explain the internal workings of RF but aims to provide explanations to decisions for a better understanding by physicians. The framework consists of two RF models. The first RF performs multi-classification to categorize individuals as normal, MCI, or AD. The second RF model only comes into play in the case of MCI, to classify whether the MCI is stable (sMCI) or progressive (pMCI). The authors used FreeSurfer software to automatically label areas of the structural MRI scans in order to provide natural language explanations using the Fuzzy Unordered Rule Induction Algorithm (FURIA) proposed by \cite{huhn2009furia}. As a result, it produces a compact set of If-THEN statements that are understandable by physicians. SHapley Additive exPlanations (SHAP) \cite{lundberg2017unified} was employed to show local/global feature contributions to final decisions of the RF models. Finally, they used a visualization tool called RF explainer on the tree decisions to provide explanation about individual modalities.

Khare et al. \cite{khare2023adazd} used electroencephalogram (EEG) which is lower-cost and less prone to radiation, as the only modality to perform binary classification of AD. After channel and feature analysis to extract the most important channels and features, the authors used the explainable boosting machine (XBM) model with three model-agnostic explainers: SHAP \cite{lundberg2017unified}, Local Interpretable Model-agnostic Explanations (LIME) \cite{ribeiro2016should}, and Morris Sensitivity (MS) \cite{morris1991factorial}. The study presented topographic maps to elucidate feature importance in terms of EEG channels. Although XBM has shown promising performance, the used dataset is small insofar as the results are not affirmative enough to support EEG signals as a standalone diagnostic tool for AD. In fact, analyzing EEG is a challenging task \cite{mustafa2023brain}, and it may not provide comprehensive information on a large scale. Nevertheless, we note that integrating EEG into a multimodal setup could be a valuable complementary aid.

Zhang et al. \cite{zhang2021explainable} proposed a 3D explainable residual attention network (3D ResAttNet) which is a deep convolutional neural network (CNN) with the addition of self-attention residual blocks and Gradient-weighted Class Activation Mapping (Grad-CAM) \cite{selvaraju2017grad}. The residual mechanism alleviates vanishing gradients in deep networks, while self-attention learns long-range dependencies. Grad-CAM is used to pinpoint relevant areas (regions associated with disease presence) in each brain scan, by calculating the gradient of the probabilities of those areas with respect to the activation of a particular unit located at a certain position in the last convolutional layer of the network. This gradient represents how sensitive the predicted probabilities are to changes in the activation of that unit, thus highlighting the contribution of each brain area to the model's decision. The authors used structural MRI (sMRI) as the single modality for two separate {\em binary} classifications: 1) AD vs CN and 2) pMCI vs sMCI.

Similarly, Yu et al. \cite{yu2022novel} used only sMRI too to perform binary diagnosis of AD (CN vs AD). Their objective was to create higher-resolution brain heatmaps to capture fine-grained details. To that end, they developed a network named MAXNet that consists of a Dual Attention Module (DAM) and a Multi-resolution Fusion Module (MFM), which learn representations that contain information at the voxel level. Additionally, they introduced High-resolution Activation Mapping (HAM) as a visualization method to enhance the quality of the heatmap. Although the algorithm can identify precise small regions in terms of voxels, validation cannot be provided as to whether the algorithm's predictions are actually \textit{correct for the correct cues}. 

Mulyadi et al. \cite{mulyadi2023estimating} tackled this issue by developing a method called eXplainable AD Likelihood Map Estimation (XADLiME), based on clinically-guided prototype learning. They measured the similarity between those prototypes and the latent features of clinical information, a clinical label, MMSE score, and age, and thereby created a \textit{pseudo} likelihood map representing the likelihood of AD across different stages. The AD likelihood map was estimated from sMRI using a feature extractor network and a reference map for comparison was obtained by a neural network with a sigmoid activation function. The estimated likelihood map can be viewed from both clinical and morphological perspectives to interpret predictions as a diagnostic tool, to help understand the likelihood of AD progression based on sMRI imaging.

While most studies, including \cite{khare2023adazd,zhang2021explainable,yu2022novel,mulyadi2023estimating}, utilize a single modality, our approach leverages multiple modalities for interpretable AD diagnosis and achieves enhanced performance. A perhaps more important differentiator of our work lies in our interpretation approach, which is rooted in our JSM framework, leading to more trustworthy medical diagnoses.

\section{Methods}
\subsection{Jacobian Saliency Map (JSM)}
Jacobian Saliency Maps (JSM) emerged recently \cite{abbas2023transformed} as a highly effective tool for deciphering the decision-making mechanisms of a deep learning model. It accomplishes this goal by defining specific zones within an input image and measuring their volumetric changes, thereby providing a precise understanding of feature attribution. Feature attribution, which ascribes significance and influence to individual features, enables us to identify the particular aspects of an input that exert significant influence on the model's output. Thus, JSM transforms data in a way that aligns with human intuition and enhances the model's interpretability before the actual deep learning pipeline, making it a promising choice for a diagnosis model debugger.

Our approach provides interpretation by computing the gradients of input with respect to \textit{weighted} elements of the input image and optimizing them toward matching the patterns of deformations highlighted by the JSM. On manipulating the input gradients, we can explore two directions: enhance their significance in relevant brain areas or reduce their significance in irrelevant brain areas. However, because the appropriate magnitudes of gradients in relevant areas are typically unknown a priori, we choose to dampen the gradients in irrelevant areas as the normal regions are a reliable reference.

We are inspired by the work done by Ross et al. \cite{ross2017right} who introduced a method to regularize a model's gradients with respect to input features based on a binary mask annotation matrix.  However, the annotation term added to the model in \cite{ross2017right} is a simple binary mask, which fails to capture correlations between different regions of a brain scan. This poses a significant limitation to the effectiveness because such correlations carry crucial diagnostic information. Additionally, the datasets used in their validation were small and synthetic, thus leaving considerable doubt on whether the method would perform well in real-world medical applications. 

Moreover, we do not use the annotation matrix but a special weighted saliency map instead. The rationale is that we prefer a matrix to not only represent the varying degrees of sensitivity of each feature to the diagnosis accurately but also highlight brain regions that exhibit deformations; in the meantime, the normal regions have to be preserved for contrast purposes. These properties cannot be achieved by the binary annotation matrix. 
In addition, by using the special saliency map to characterize the deformations in the brain, it also allows us to align the constraints on gradients (a penalty we formulate later in \eqref{eq:loss}, \eqref{eq:spline}) with the domain-specific knowledge related to the medical problem. We perform both registration and Jacobian using Advanced Normalization Tools (ANTs) \cite{avants2009advanced}.

{\bf Intuition behind JSM.} Ideally, the presence of a disease can be identified by comparing an individual's many scans over a long period of time. Since this is usually not feasible in practice, we can compare an individual's scan to a standardized healthy brain template to assess its local changes using a deformation map. JSM is derived from non-linear image registration, which is a process designed to both minimize variations among individual subjects and align all different images with a standardized template. JSM then examines the deformations that transform the anatomical structures of individuals onto a common standard space, through which we can deduce the relative volume differences between each individual and the template. This analysis aids in pinpointing statistically significant anatomical variations across diverse populations, such as distinguishing between AD patients and healthy elderly individuals. In our investigation, we utilized the Montreal Neurological Institute's 152 brain template (MNI152) as the reference. This process of image registration and JSM computation can be attributed to the realm of computational anatomy, specifically recognized as tensor-based morphometry \cite{riyahi2018quantifying}, and has not been adequately explored in the context of dementia.

{\bf Formulation of JSM.} In medical image processing, computing deformation involves first aligning a source image $M$ with a target image $F$. This is achieved by using a transformation $\phi$ that maps points in $M$ to their corresponding points in $F$. Then, the displacement between these corresponding points is represented as a Deformation Vector Field:
\begin{equation}
    \vec{v}(x, y, z)= \phi(x, y, z)-(x, y, z).
\end{equation}

To transform a point $(x,y,z)$ to $\phi(x,y,z)$, it is necessary to impose a regularization constraint in order to ensure that the deformation is seamless, one-to-one, and differentiable. This is framed as an optimization problem that minimizes the following cost function:
%\red{is this correct? i feel it should be to {\em maximize} $L_{sim}(\phi(M), F)$ instead (or minimize the minus of L)? if so, then MI needs to be corrected too}\blue{Yes. Since it's a loss function that we want to minimize. The first term is conventionally referred to as Lsim for similarity but it actually penalizesdifferences in appearance between M and F.}\red{still not corrected}

\begin{equation}\label{eq:loss}
    L(\phi, M, F) = -L_{sim}(\phi(M),F) + \alpha L_{Reg}(\phi)
\end{equation}
where $L_{sim}$ is a similarity measure between two images, the transformed image $\phi(M)$ and $F$, and $L_{Reg}$ is a regularization term that enforces the desired properties on the deformation.
%\red{what does $\circ$ mean?}\blue{Transforming M to F}\red{still not corrected}
We use Mattes Mutual Information (MI) \cite{mattes2003pet} as our similarity measure, i.e.,
\begin{equation}
L_{sim} = MI(M', F) = \sum_{m'} \sum_{f} P(m', f) \log\left(\frac{P(m', f)}{Q_1(m')Q_2(f)}\right)
\end{equation}
where $M'$ denotes $\phi(M)$ for simplicity. $P(m', f)$ is the joint probability distribution of the intensity of voxel $m'$ in image $M'$ and that of voxel $f$ in image $F$, and $Q_1(m')$ and $Q_2(f)$ represent the marginal probability distributions of $m'$ and $f$, respectively. For the regularizer, we use the B-spline regularization from \cite{tustison2013explicit}:
\begin{equation}\label{eq:spline}
L_{Reg} = L_{\text{B-spline}} = \int \left| \nabla^2 \phi(x, y, z) \right|^2 dV
\end{equation}
where $\nabla^2$ denotes the second-order derivative, $dV$ represent seach voxel $(x, y, z)$, and we integrate over the entire spatial domain.

After solving for the transformation $\phi$, we compute a Jacobian matrix $J$ from the deformation vector field $\vec{v}$, by calculating the first derivative of $\vec{v}$ at each voxel to encode local deformations including stretching, shearing, and rotation. That is,
\begin{equation}
J(v)=
\begin{bmatrix}
\frac{\partial v_x}{\partial x} & \frac{\partial v_x}{\partial y} & \frac{\partial v_x}{\partial z} \\
\frac{\partial v_y}{\partial x} & \frac{\partial v_y}{\partial y} & \frac{\partial v_y}{\partial z} \\
\frac{\partial v_z}{\partial x} & \frac{\partial v_z}{\partial y} & \frac{\partial v_z}{\partial z}
\end{bmatrix}
\end{equation}
Then, denoting the Jacobian determinant by $Det(J)$, we calculate it for every voxel $v(x,y,z)$ as
\begin{align}
\Resize{12.5cm}{Det(J) = \, \frac{\partial v_x}{\partial x} \left(\frac{\partial v_y}{\partial y} \frac{\partial v_z}{\partial z} - \frac{\partial v_y}{\partial z} \frac{\partial v_z}{\partial y}\right) - \frac{\partial v_x}{\partial y} \left(\frac{\partial v_y}{\partial x} \frac{\partial v_z}{\partial z} - \frac{\partial v_y}{\partial z} \frac{\partial v_z}{\partial x}\right) + \frac{\partial v_x}{\partial z} \left(\frac{\partial v_y}{\partial x} \frac{\partial v_z}{\partial y} - \frac{\partial v_y}{\partial y} \frac{\partial v_z}{\partial x}\right)}
\end{align}
which forms what we call a \textit{Jacobian Saliency Map} $JSM$ of the source image $M$:
\begin{equation}
\begin{array}{l}
JSM(M)= 
\begin{bmatrix}
& \vdots &\\
\dots & Det(J(v(x,y,z)) & \dots \\
& \vdots &\\
\end{bmatrix}_{\begin{matrix}x=1...W{\rm\ (width)} \\ y=1...H {\rm\ (height)} \\ z=1...D {\rm\ \ (depth)} \end{matrix}} \\
\\
\hspace{-0.4cm}\text{At each voxel: }
\begin{cases}
\text{{volume expansion}} & \text{{if }} Det(J) > 1 \\
\text{{no change}} & \text{{if }} Det(J) = 1 \\
\text{{volume compression}} & \text{{if }} Det(J) < 1 
\end{cases}
\end{array}\label{volume_changes}
\end{equation}
%W, H, and D represent the width, height, and depth of the 3D medical image.

Volumetric changes refer to the alteration in volume at the level of individual voxels within a medical image. This JSM helps us to identify the volumetric ratio of the brain image at the voxel level before and after transformation $\phi$, which indicates the brain's volume change.

\subsection{Jacobian-Augmented Loss Function (JAL)} 
By breaking down an input image into distinct regions and measuring how they are transformed, JSM provides precise insight into the complexities of feature attribution and thus model explainability. To this end, we take an innovative approach to explore the possibility of leveraging JSM as a powerful {\em model debugger}, for which we incorporate the JSM formulated above into the loss function $\mathcal{L}$ of a medical diagnostic model:
\begin{align}\label{jsm_eqn}
\mathcal{L}( \boldsymbol{x}, \boldsymbol{y}, JSM) &= - \sum_{k=1}^K y_{k} \log (\hat{y}_{k})+ \lambda \sum_{d=1}^D \sum_{p=1}^P \left( w_{dp}JSM_{dp} \frac{\partial}{\partial x_{dp}} \sum_{k=1}^K \log (\hat{y}_{n k})\right)^2 
\end{align}

Given input data $\boldsymbol{x}$ and data label $\boldsymbol{y}$, the loss function $\mathcal{L}(\lambda, \boldsymbol{x}, \boldsymbol{y}, JSM)$ consists of the training loss (first term) and a novel, JSM-based regularization term (second term). It introduces an emphasis on the importance of anatomical changes captured by the Jacobian map values. Specifically, $JSM_{dp}$ represents the JSM values associated with the $d^{th}$ feature and $p^{th}$ spatial dimension (width, height, and depth in 3D), $log(\hat{y}_{k})$ represents the natural logarithm of the predicted probability of $x$ belonging to class $k$, and $\frac{\partial}{\partial x_{dp}}$ reflects the partial derivative with respect to the $d^{th}$ feature of $x$ in the $p^{th}$ spatial dimension. Thus, this regularization term aims to not only enhance interpretability but also rectify predictions by mitigating the influence of irrelevant cues.

With reference to Equation \eqref{volume_changes}, we add a weight matrix $W$ to give more importance to areas in the JSM that have volumetric changes (expansion or compression) and discourage the input gradients from being significant in areas with no volumetric changes (marked by 1). Hence each element of $W$ is defined as follows:
\begin{equation}
w_{dp} = 
\begin{cases}
  \text{{feature\_weight}}, & \text{{if }} JSM_{dp} \neq 1 \\
  \text{{debug\_weight}}, & \text{{otherwise}}
\end{cases}
\end{equation}
Both feature\_weight and debug\_weight are hyperparameters that indicate the level of importance in every region. We designate a debug weight of 0.2. This choice is deliberate, allowing us to down-weight potentially misleading features while still retaining them as a reference for contrasting relevant features. On the other hand, since regions of volumetric changes are of higher importance, we assign them a feature weight of 0.8. Hence, we debug the model through weighting features according to its relevance. Note that we don't eliminate irrelevant areas by giving a weight of zero to preserve correlations in the brain volume. More rigorous hyperparameter tuning of such weights can be incorporated in future work.
\begin{figure*}[!t]
  \centering
 \includegraphics[width=0.7\linewidth]{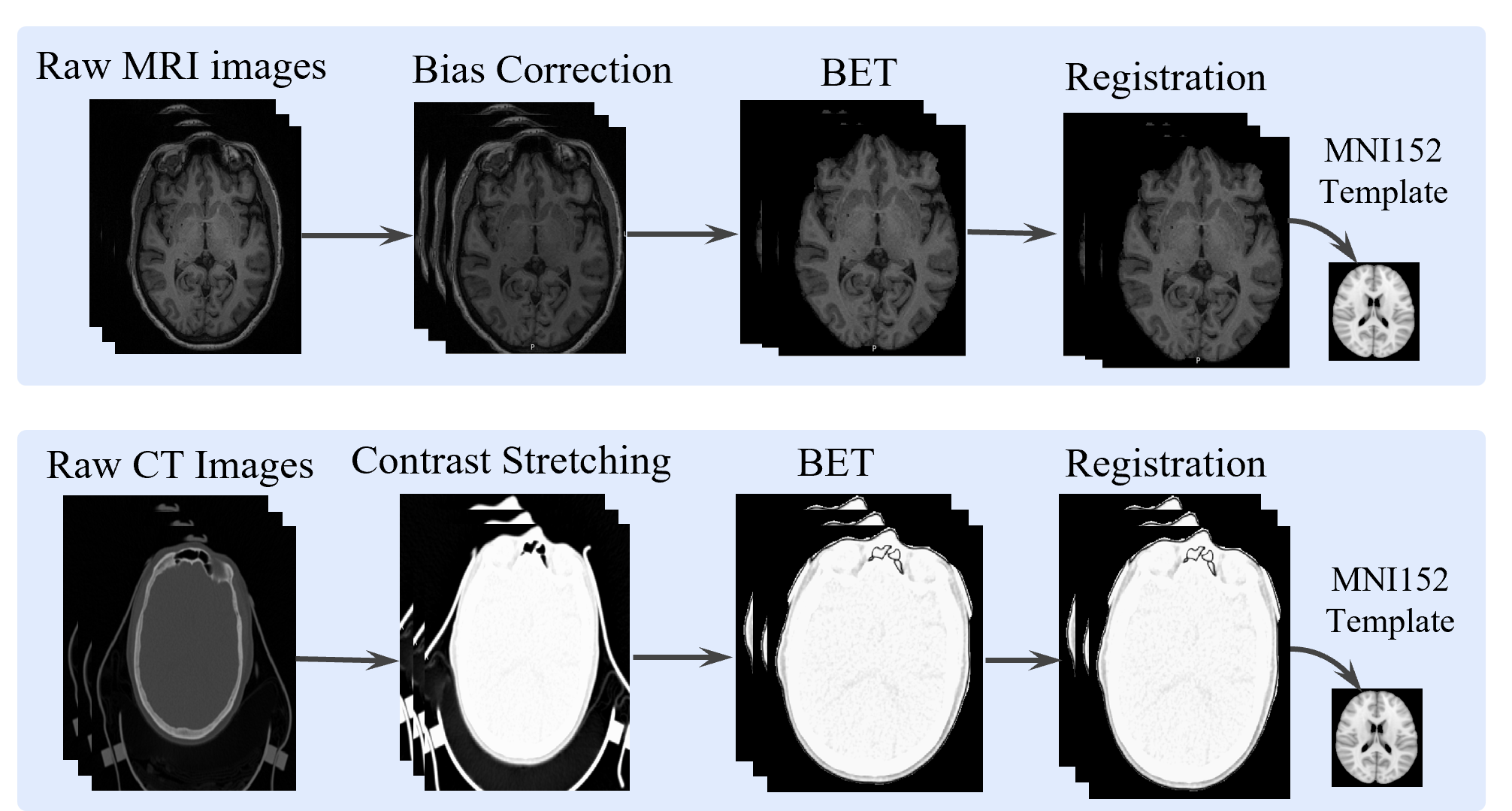}
  \caption{Preprocessing pipelines for MRI and CT scans, involving bias field correction for MRI, contrast stretching for CT to enhance diagnostic values, BET for brain extraction, and registering CT and MRI to MNI152 brain template.}
  \label{preprocessing}\vspace{-0.4cm}
\end{figure*}
\section{Experiments}
\subsection{Dataset}
In our experiments, we utilized the recently published OASIS-3 dataset. This dataset includes multi-modality data such as MRI, PET, and CT scan images from a diverse group of 1377 participants. Among these participants, 755 were cognitively normal (CN) adults, while the remaining 622 individuals showed different levels of cognitive decline. The age range of the participants was extensive, spanning from 42 to 95 years. CT imaging was used to detect whether certain areas of the brain were shrinking, which can be an indication of  AD. On the other hand, MRI provided detailed images of the body and a clear view of progressive cerebral atrophy, which is most visible through T1-weighted volumetric sequences. Consideration of PET data was deferred due to its temporal nature, making it more suitable for future spatiotemporal analyses.

During clinical assessments and diagnoses, the clinical dementia rating (CDR) scores of the participants were utilized. The scores range from 0 to 3, with 0 indicating no AD dementia and 3 indicating severe AD dementia. The very mild stage (rating 0.5) of dementia is similar to the Mild Cognitive Impairment (MCI) stage of AD. As mentioned before, we combined moderate and severe AD due to the very low number of subjects with severe AD. Ultimately, we created four classes based on CDR scores: normal, MCI, mild AD, and severe AD. Having the same subjects in multiple sets (train and test sets) can result in the model overfitting to those individuals, potentially leading to a subpar performance on new, unseen subjects. Hence, for patients who underwent multiple sessions, we preserved the initial MRI session and selected the CT session that was closest in date to that MRI.  

\subsection{Preprocessing}
Our preprocessing pipelines for MRI and CT scans are shown in \fref{preprocessing}. To minimize any spatially varying intensity bias that may result from factors such as magnetic field inhomogeneities and acquisition artifacts, we use the FMRIB's Linear Image Registration Tool (FLIRT) \cite{jenkinson2002improved} for bias field correction on MRI images. Meanwhile, we utilize a technique called contrast stretching on CT images to enhance their diagnostic value and visual perception. This process involves adjusting the pixel intensities to fully utilize the display's dynamic range. For CT images, we followed the framework established by Kuijf et al. \cite{kuijf2013registration}. Also, we apply the Brain Extraction Tool (BET) \cite{smith2002fast} to eliminate non-brain portions from both MRI and CT images. Finally, we register both CT and MRI to the  MNI152 brain template. Brain templates are typically created for MRI images, making it difficult to register a CT image to an MRI template. We overcame this by adhering to a method in \cite{kuijf2013registration} which involves identifying corresponding landmarks in the CT image and MRI template and then using these landmarks to align the images.

\subsection{Multimodal Classification} 
\begin{figure}[!t]
  \centering
 \includegraphics[width=\linewidth]{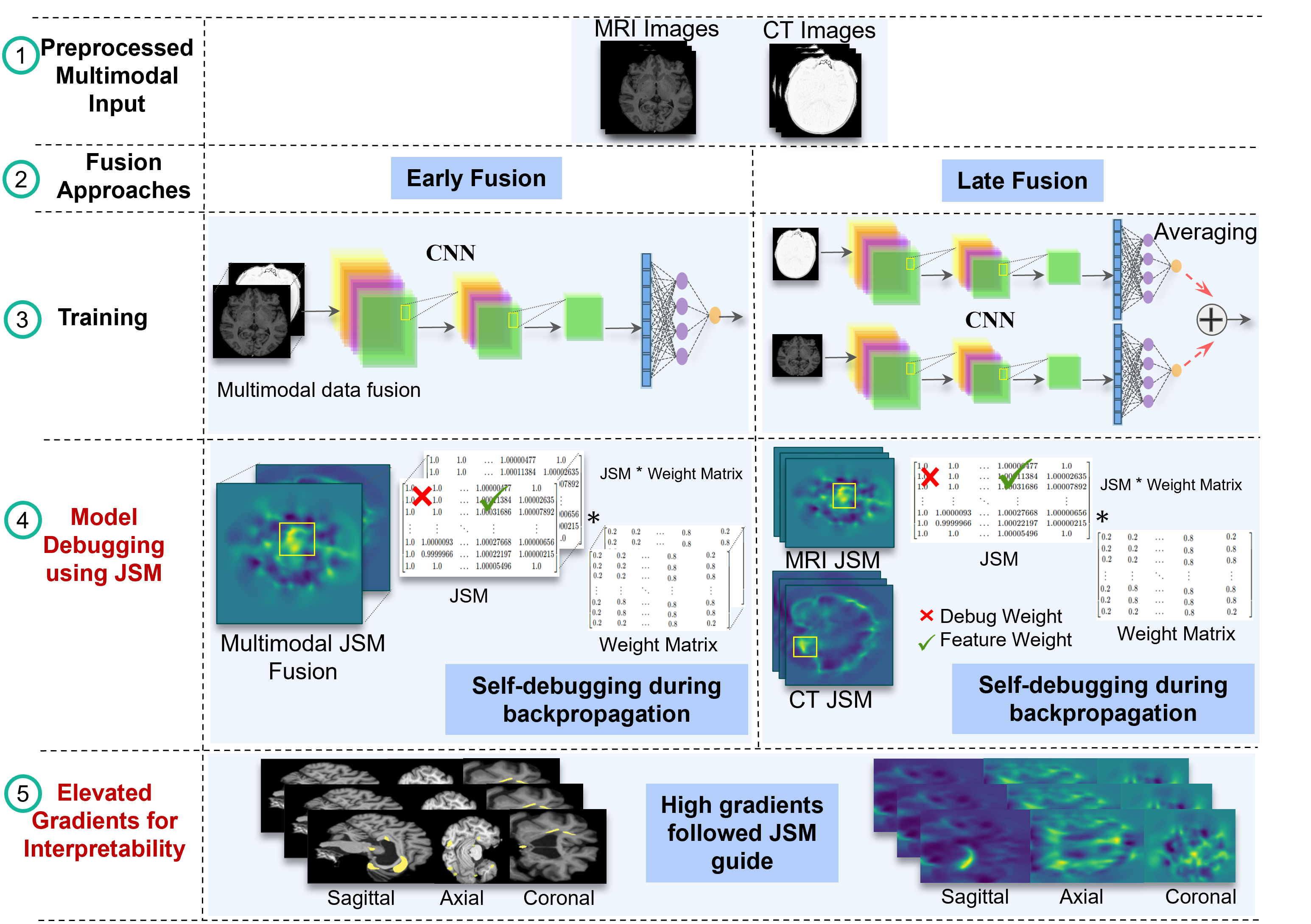}%Images/model4.png\textbf{}
  \caption{The complete pipeline. Model debugging using JSM is integrated into training and takes effect during backpropagation, for each modality (late fusion) or both (early fusion). Final predictions are interpreted by plotting elevated gradients overlaid on input images.}
  \label{model}
\end{figure}

Medical images, especially those related to the brain, are typically 3D. This introduces a computational burden into complex deep neural networks. Our approach seeks to harness the guidance provided by the JSM, and by incorporating such insights from JSM, we aim to develop lighter convolutional neural networks (CNNs) that alleviate computational burdens without compromising model performance. In view of the multimodal nature of our study, we incorporated two data fusion techniques: 1) Late Fusion and 2) Early Fusion. As shown in \fref{model}, late fusion adopts a dual-branch structure, treating each modality independently and subsequently aggregating their predictions through an averaging mechanism. This approach is particularly advantageous for JAL, where debugging is performed separately for each modality through its own JSM. On the other hand, early fusion involves concatenating the input images as well as their corresponding JSM maps, which allows the model to glean correlations between the two modalities and concurrently debug predictions for the input holistically.
\begin{wraptable}{R}{0.5\linewidth} \vspace{-1.3\baselineskip}
    \centering
    \caption{CNN Architecture for both branches}
    \label{architecture_details}
    \setlength{\tabcolsep}{1.2em}
    \renewcommand{\arraystretch}{1}
    \resizebox{\linewidth}{!}{%
        \small
        \begin{tabular}{p{1.6cm}p{2.88cm}p{3.3cm}}
            \hline\hline
            \textbf{Layers} & \textbf{Parameters} & \textbf{Output Size} \\
            \midrule
            Input & Batch Size 10 & $10\times1\times182\times256\times512$ \\
            \midrule
            \multirow{3}{*}{Conv1} & Stride 1 & \multirow{3}{*}{$10\times 4 \times 182\times256\times512$} \\
            & Padding 1 & \\
            & Kernel Size $3\times3\times3$ & \\
            \midrule
            BatchNorm1 & Momentum=0.9 & $10\times4\times182\times256\times512$ \\
            \midrule
            Dropout1 & Dropout rate 0.5 & $10\times4\times182\times256\times512$ \\
            \midrule
            \multirow{2}{*}{MaxPool1} & Stride 2 & \multirow{2}{*}{$10\times4\times91\times128\times256$} \\
            & Kernel Size $2\times2\times 2$ & \\
            \midrule
            \multirow{3}{*}{Conv2} & Stride 1 & \multirow{3}{*}{$10\times8\times91\times128\times256$} \\
            & Padding 1 & \\
            & Kernel Size $3\times3\times3$ & \\
            \midrule
            BatchNorm2 & Momentum=0.9 &$10\times8\times91\times128\times256$ \\
            \midrule
            Dropout2 & Dropout rate 0.2 & $10\times8\times91\times128\times 256$ \\
            \midrule
            \multirow{2}{*}{MaxPool2} & Stride 2 & \multirow{2}{*}{$10 \times 8 \times 45\times64\times128$} \\
            & Kernel Size $2\times2\times2$ & \\
            \midrule
            Flatten & & $10\times2949120$ \\
            \midrule
            {{Full-Conn.}} & &$10\times4$ \\
            \hline\hline
        \end{tabular}} 
        %\vspace{-4cm}
\end{wraptable}

Our lightweight CNN model \cite{mustafa2023diagnosing} contains two convolutional layers coupled with batch normalization, ReLU activation, dropout, and max-pooling operations, which capture spatial hierarchies and correlation patterns in the input data. The convolutional layers use a kernel size of 3x3x3, stride of 1, and padding, with a dropout of rate 0.2 for regularization. The max-pooling operation reduces spatial dimensions to 2x2x2.  The 3D tensor is reshaped into a 1D tensor by a flattening layer, preparing it for the subsequent fully connected subnetwork, which performs the overall feature integration and classification. All the specifications are presented in Table \ref{architecture_details}. Finally, a softmax layer converts the predicted scores into probabilities. In the late fusion setup, the final prediction is computed by averaging the probabilities from both branches.
\vspace{-0.4cm}
\subsection{Performance Evaluation}
 % Adjust the space as needed
%\begin{wraptable}{R}[0pt]{0.4\textwidth}\vspace{-1\baselineskip}
%\raggedleft

%\vspace{1em} % Adjust the space as needed

We trained our model on a 40GB A100 GPU with batch size 10. Our model was able to quickly converge within only 20 epochs. To address the class imbalance problem in AD, we utilized the Adaptive Synthetic (ADASYN) \cite{he2008adasyn} oversampling algorithm to generate synthetic samples for minority classes during training. Another challenge is that OASIS-3 included identical subjects from multiple sessions across training and test sets may contribute to overfitting, as the model can become excessively attuned to the characteristics of those particular subjects and sessions \cite{altay2021preclinical}. To address this, we take meticulous care to ensure that only subjects from distinct sessions are grouped within either the training set or the test set (but not both).

Table \ref{comparison} summarizes the performance comparison in terms of accuracy, sensitivity, and specificity between our approach and the state-of-the-art. Please note that while \textit{precision} and \textit{recall} are more general terms used in machine learning, \textit{sensitivity} and \textit{specificity} are often preferred in medical and diagnostic fields due to their direct interpretation in the context of disease detection and diagnosis.
Table \ref{comparison} shows that our testing accuracy across four classes surpasses all the baselines that employ the same dataset. Massalimova et al. \cite{massalimova2021input} achieved marginally higher sensitivity and specificity, but it is crucial to note that our model handles a larger number of classes, making it more challenging to achieve higher accuracy. In addition, 
\cite{massalimova2021input} uses ResNet18 while our model is significantly lighter. In fact, our model is lighter than nearly all the baselines. Basheer et al. \cite{basheer2021computational} used features like CDR, MMSE, age, gender, etc along with MRI images, and found that age and gender had substantial positive impact on performance. 
In our case, we achieved superior performance solely with images as we aimed to test our model using spatial features.

\begin{figure*}[!t]\label{cnn_perf}
     \begin{subfigure}[b]{0.34\textwidth}
         \centering
         \includegraphics[width=\linewidth]{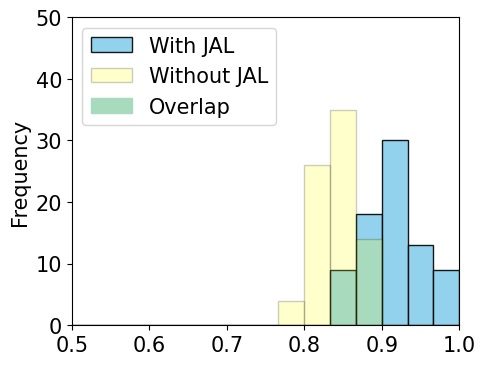}
\caption{\centering Accuracy}
     \end{subfigure}
     \hfill
\begin{subfigure}[b]{0.32\textwidth}
         \centering
        \includegraphics[width=\linewidth]{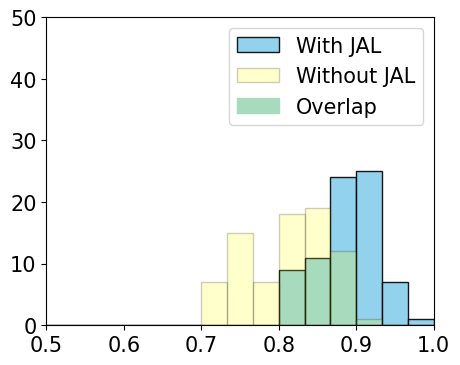}
   \caption{\centering  Sensitivity}
     \end{subfigure}
     \hfill
     \begin{subfigure}[b]{0.32\textwidth}
         \centering
        \includegraphics[width=\linewidth]{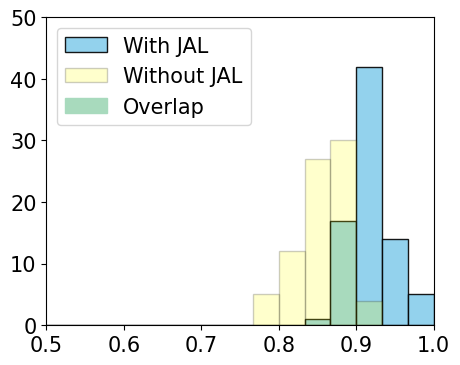}
   \caption{\centering  Specificity}
     \end{subfigure}

     %-------
\begin{subfigure}[b]{0.34\textwidth}
         \centering
         \includegraphics[width=\linewidth]{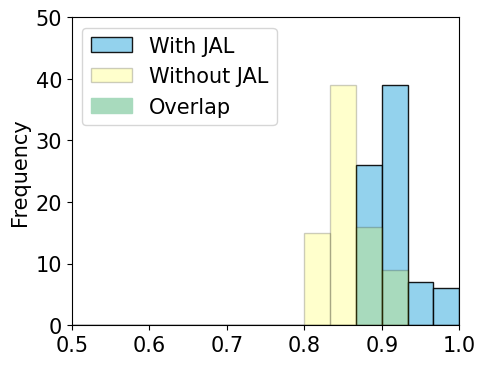}
\caption{\centering Accuracy}
     \end{subfigure}
\hfill
\begin{subfigure}[b]{0.32\textwidth}
         \centering
        \includegraphics[width=\linewidth]{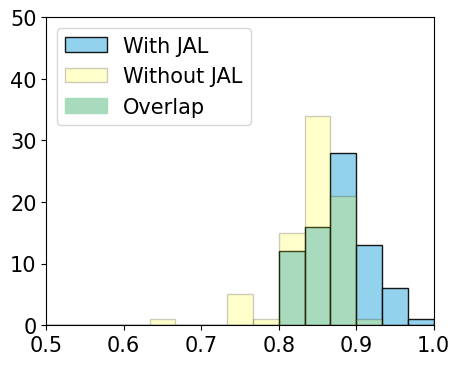}
   \caption{\centering  Sensitivity}
     \end{subfigure}
     \hfill
     \begin{subfigure}[b]{0.32\textwidth}
         \centering
        \includegraphics[width=\linewidth]{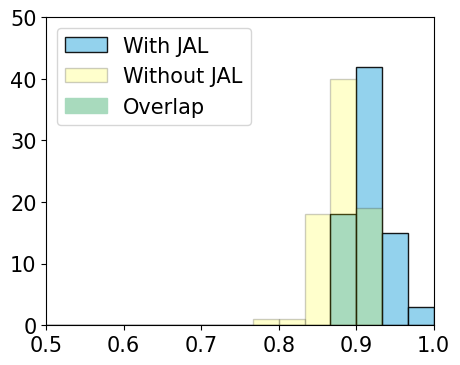}
   \caption{\centering  Specificity}
     \end{subfigure}
 \caption{Ablation study on JAL in terms of performance histograms. (a-c): Early fusion, (d-f): Late fusion.}
   \label{histogram_test}
\end{figure*}

{\bf Ablation Study.}
To provide an in-depth assessment of the impact of JAL on our model's performance, we conducted an ablation study on models with and without JAL. This was achieved by setting the JSM term in \eqref{jsm_eqn} to zero. The results are presented in \fref{histogram_test}, which provides histograms of model performance distribution across multiple mini-batches in the test set. The overlap represents the intersection between the \textit{with JAL} and \textit{without JAL} conditions, indicating the extent to which the model's performance remains consistent regardless of the presence or absence of JAL. The histogram demonstrates that the model performance significantly improves in terms of all the metrics (accuracy, sensitivity, and specificity) when JAL is incorporated, as evidenced by the rightward shift of the distributions in the blue bars in comparison with the yellow bars.  This compellingly demonstrates the impact of incorporating JAL on the model's decisions.
For a more comprehensive evaluation, \tref{ablation} dissects the four classes and provides more detailed results. It shows substantial improvements over all the AD stages (CN, MCI, MLD, SEV), further affirming the efficacy of JAL. \tref{ablation} also allows us to see that performance for CN and SEV (severe) are relatively higher than MCI and MLD, which is because MCI and MLD have more subtle differences in dementia patterns, making them more intricate to discern. Nevertheless, our model with JAL exhibits evenly promising outcomes for all stages. Scores in \tref{comparison} are the macro average of the scores in \tref{ablation}. 

\begin{table*}[!t]
\centering
\caption{Comparison with reported state-of-the-art using OASIS-3 dataset for AD classification}
\resizebox{.9\linewidth}{!}{%
\small
\begin{tblr}{|p{3.15 cm}p{2.5cm}p{3.1cm} p{2.42cm}p{2.38cm}p{2.25cm}| }
\hline
\textbf{Model}  & \textbf{Modalities}  & \textbf{Classes}  & \textbf{Sensitivity (\%)} & \textbf{Specificity (\%)} & \textbf{Accuracy (\%)} \\ 
\hline\hline
Salami et al. \cite{salami2022designing}           & MRI          & AD, CN        & \centering 86.01    & \centering85.04       & \centering87.75 \\
\hline
Massalimova et al. \cite{massalimova2021input}    & MRI      & CN, MCI, AD   & \centering96      & \centering 96         & \centering96  \\
\hline
Lazli et al. \cite{lazli2019computer}             & MRI, PET           & AD, CN   & \centering 92.00    & \centering 91.78      & \centering91.46  \\
\hline
Basheer et al. \cite{basheer2021computational} & MRI, features &   AD, CN  & \centering82.3 & \centering *NP & \centering92.3\\\hline
 Castellano et al. \cite{castellano2021detection} & PET &   AD, CN  & \centering NP & \centering NP & \centering80\\
%Basheer et al. \cite{basheer2021computational} & MRI &   AD, CN  &  & &\\
\hline\hline
%Yasmine et al. \cite{mustafa2023diagnosing} & MRI, CT  &  CN, MCI, MOD, SEV   & \centering 97.19 & \centering 95.19  & \centering 98.76 \\\hline
\SetCell[r=2]{l, font=\bfseries} Our work & MRI, CT (Early) & CN, MCI, MOD, SEV & \centering  92.72 &  \centering  95 & \centering 91.31 \\
\hline
& MRI, CT (Late) & CN, MCI, MOD, SEV & \centering  93.5 & \centering  93.5 & \centering 95.37\\
\hline
\SetCell[r=1]{l, font=\small}*NP: Not Provided \\
\hline
\end{tblr}}
\label{comparison}
\end{table*}

\begin{table*}
\setlength{\tabcolsep}{.4em}
\centering
\renewcommand{\arraystretch}{1.3}
\caption{Ablation study comparing model performance with and without JAL in Late and Early Fusion setups.}
\resizebox{.9\linewidth}{!}{%
\begin{tabular}{|p{0.3 cm}|p{1.7 cm}|p{1cm}p{1cm} p{1cm}p{1cm}|p{1cm}p{1 cm}p{1cm}p{1cm} |p{1cm}p{1cm}p{1cm}p{1cm}|}
\hline
\multirow{2.2}{1mm}{\rotatebox[origin=p]{90}{{\footnotesize{ Fusion}}}}&\centering Loss & \multicolumn{4}{c|}{Sensitivity (\%)} & \multicolumn{4}{c|}{Specificity (\%)} & \multicolumn{4}{c|}{Accuracy (\%)} \\
\cline{3-5} \cline{6-9} \cline{10-14}
&\centering Function&{CN} & {MCI} &  {MLD} & {SEV} & {CN} & {MCI} & {MLD} &{SEV} & {CN} & {MCI} &  {MLD} & {SEV} \\ \hline\hline
\multirow{2}{1mm}{\rotatebox[origin=p]{90}{\centering {\footnotesize{Early}}}}
&\textbf{w/o JAL}  & 80.34 & 78.96 & 74.3 & 80.96     & 84.34 & 89.5 & 84.34 &89    & 80 & 84.3 & 80.21 & 89  \\
\cline{2-14}
& \cellcolor{blue!10}\textbf{w/ JAL}     & \cellcolor{blue!10}90.3 &  \cellcolor{blue!10}94.6& \cellcolor{blue!10}94 & \cellcolor{blue!10}92     & \cellcolor{blue!10}99.12  & \cellcolor{blue!10}92.01 & \cellcolor{blue!10}99 & \cellcolor{blue!10}90  & \cellcolor{blue!10}98.8& \cellcolor{blue!10}87.33 & \cellcolor{blue!10}86.59 & \cellcolor{blue!10}92.5 \\ \hline\hline
\multirow{2}{1mm}{\rotatebox[origin=p]{90}{\centering {\footnotesize{Late}}}}&\textbf{w/o JAL}    & 87.7 & 87.9 & 85.5 & 87.6   & 87.8 & 86.6 & 86.6 & 85.2  & 88.8 & 86.1 & 88.4 & 87.3\\
\cline{2-14}
& \cellcolor{blue!10}\textbf{w/ JAL}    & \cellcolor{blue!10}95.3 & \cellcolor{blue!10}93.3 & \cellcolor{blue!10}93.3 & \cellcolor{blue!10}91.5 &  \cellcolor{blue!10}99.6 & \cellcolor{blue!10}92 & \cellcolor{blue!10}92 & \cellcolor{blue!10}92.2 &  \cellcolor{blue!10}99.8 & \cellcolor{blue!10}92 & \cellcolor{blue!10}93 & \cellcolor{blue!10}96.7   \\ \hline
\end{tabular}\label{ablation}}
\end{table*}

\begin{figure*}[!h]
  \centering
 \includegraphics[width=\linewidth]{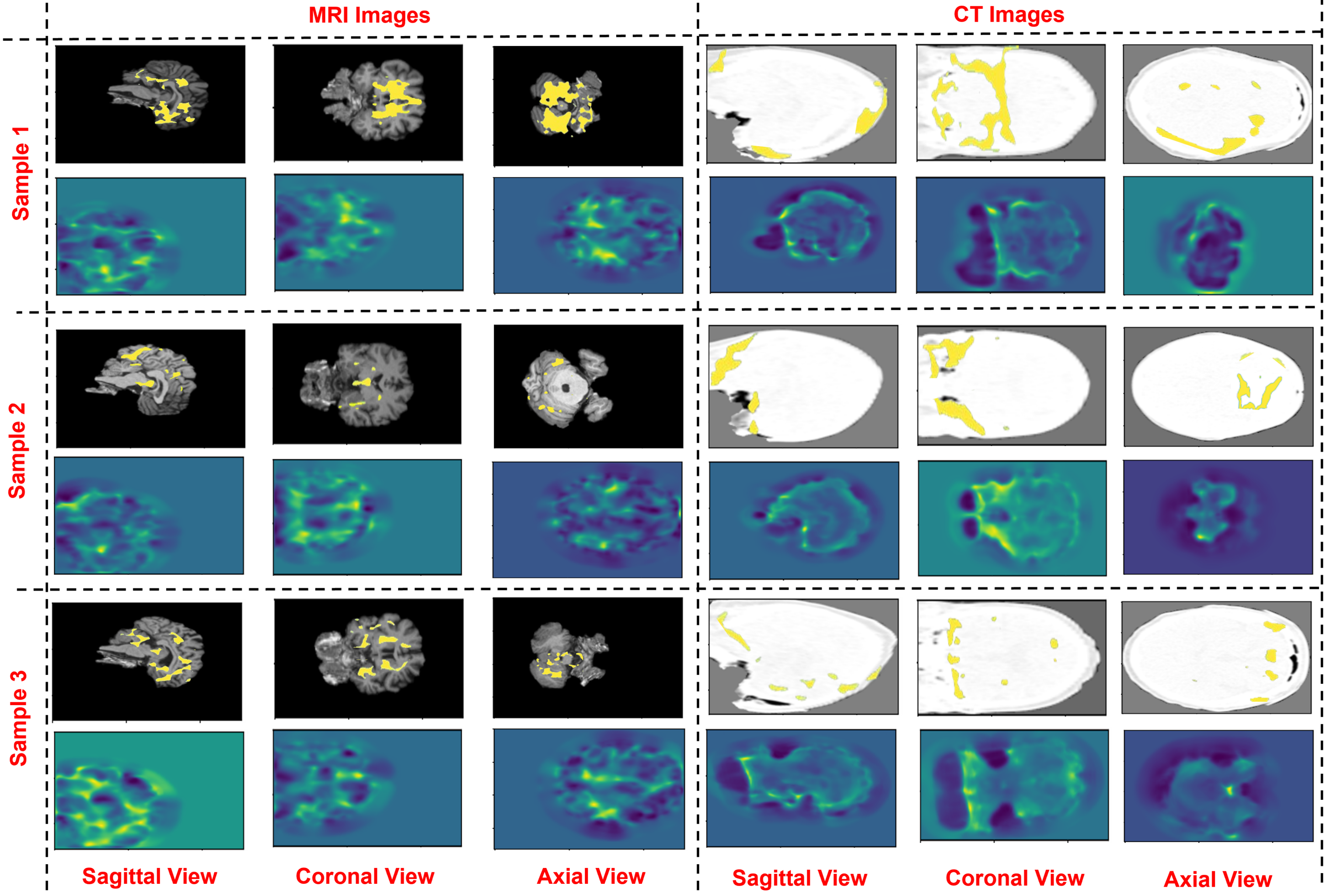}
 \caption{Visualization of larger gradients in JSM-indicated deformation areas for MRI and CT modalities.}
  \label{mri}
\end{figure*}
{\bf Interpretability.}
We examined the similarity between the volumetric changes characterized by JSM and the decision-making process of the neural network. By plotting gradients overlaid on their corresponding input images, we observed that they closely aligned with the patterns highlighted by the JSM. This can be seen from \fref{mri} which provides samples from the dataset showing the axial, sagittal, and coronal views of each MRI and CT images juxtaposed with the corresponding JSM views that highlight brain deformations. This desirable alignment is attributed to our JAL loss function which incorporates the JSM during the model debugging process, contributing to the model trustworthiness by promoting transparency.

Furthermore, incorporating JSM in JAL during model debugging also enables the model to learn and adapt to the highlighted deformations, and hence serves as a powerful tool for refining the model performance as well, fostering a more accurate and \textit{informed} decision-making process. 

\section{Conclusion}
This paper introduces a new approach to trustworthy medical diagnoses, by addressing two key challenges: model explainability and reliability. On explainability, we leverage Jacobian saliency maps (JSM) to provide informative and interpretable guide for feature learning, as well as capture subtle morphological changes associated with the disease. On reliability, we incorporate JSM into the loss function as a self-debugger to direct the model to critical (disease-relevant) regions during training, avoiding the \textit{Clever-Hans} behavior.
Our approach not only helps rectify erroneous predictions but also identifies regions in a post-hoc manner with elevated gradients for interpretability enhancement. (Note that post-hoc is for visualization only; our main approach of JAL/debugging is a during-modeling approach.) 
Our extensive evaluation underscores the success of JSM via a Jacobian-augmented loss function (JAL), leading to substantial accuracy improvement (by up to 10\%) and greater model interpretability in identifying significant brain areas that lead to diagnostic predictions. Our XAI approach also works seamlessly with our multimodal data fusion methods and provides explanation in both early and late fusion setups.
%The JSM illuminates areas of compression or expansion relative to a healthy brain template, thereby highlighting deformations in the brain.
%enhancing the understanding of the model's inner workings. Users and stakeholders are more likely to trust a model whose decision-making processes are not only accurate but also intelligible and aligned with human-understandable features.

%directing the model makes the decision-making process more transparent by basing model guidance on regions (in the case of Alzheimer's disease) that exhibit morphological changes in the brain, which are highlighted by the Jacobian saliency maps (JSM).
%\begin{figure}
%\includegraphics[width=\textwidth]{fig1.eps}
%\caption{A figure caption is always placed below the illustration.
%Please note that short captions are centered, while long ones are
%justified by the macro package automatically.} \label{fig1}
%\end{figure}

%
% the environments 'definition', 'lemma', 'proposition', 'corollary',
% 'remark', and 'example' are defined in the LLNCS documentclass as well.
%

%\subsubsection{Acknowledgements} Please place your acknowledgments at \cite{chettri2023clever}the end of the paper, preceded by an unnumbered run-in heading (i.e.3rd-level heading).

{\small
\bibliographystyle{splncs04}
\bibliography{mybibliography}}

\end{document}